\documentclass[10pt,journal,compsoc]{IEEEtran}
\ifCLASSOPTIONcompsoc
\usepackage[nocompress]{cite}
\usepackage{makecell}
\usepackage{graphicx} 
\usepackage{multirow}
\usepackage{tablefootnote}
\usepackage{threeparttable}
\else
\usepackage{cite}
\fi
\ifCLASSINFOpdf
\else
\fi
\begin{document}
\title{Learning to Rank Onset-Occurring-Offset Representations for Micro-Expression Recognition}
%
%
%
%

\author{Jie~Zhu,
Yuan~Zong$^*$,~\IEEEmembership{Member,~IEEE,}
Jingang~Shi,
Cheng~Lu,
Hongli~Chang, and
Wenming~Zheng$^*$,~\IEEEmembership{Senior~Member,~IEEE}
\IEEEcompsocitemizethanks{
\IEEEcompsocthanksitem J. Zhu and H. Chang are with the Key Laboratory of Child Development and Learning Science of Ministry of Education, Southeast University, Nanjing 210096, China and also with the School of Information Science and Engineering, Southeast University, Nanjing 210096, China.
\IEEEcompsocthanksitem Y. Zong, C. Lu and W. Zheng are with the Key Laboratory of Child Development and Learning Science of Ministry of Education, Southeast University, Nanjing 210096, China and also with the School of Biological Science and Medicial Engineering, Southeast University, Nanjing 210096, China.
\IEEEcompsocthanksitem J. Shi is with the School of Software, Xi'an Jiao Tong University, Xi\'an 710049, China.
}
\thanks{* indicates the corresponding authors.}}

%
%

\markboth{Journal of \LaTeX\ Class Files,~Vol.~14, No.~8, August~2015}%
{Shell \MakeLowercase{\textit{et al.}}: Bare Demo of IEEEtran.cls for Computer Society Journals}
%



\IEEEtitleabstractindextext{%
\begin{abstract}
This paper focuses on the research of micro-expression recognition (MER) and proposes a flexible and reliable deep learning method called learning to rank onset-occurring-offset representations (LTR3O). The LTR3O method introduces a dynamic and reduced-size sequence structure known as 3O, which consists of onset, occurring, and offset frames, for representing micro-expressions (MEs). This structure facilitates the subsequent learning of ME-discriminative features. A noteworthy advantage of the 3O structure is its flexibility, as the occurring frame is randomly extracted from the original ME sequence without the need for accurate frame spotting methods. Based on the 3O structures, LTR3O generates multiple 3O representation candidates for each ME sample and incorporates well-designed modules to measure and calibrate their emotional expressiveness. This calibration process ensures that the distribution of these candidates aligns with that of macro-expressions (MaMs) over time. Consequently, the visibility of MEs can be implicitly enhanced, facilitating the reliable learning of more discriminative features for MER. Extensive experiments were conducted to evaluate the performance of LTR3O using three widely-used ME databases: CASME II, SMIC, and SAMM. The experimental results demonstrate the effectiveness and superior performance of LTR3O, particularly in terms of its flexibility and reliability, when compared to recent state-of-the-art MER methods.
\end{abstract}

}

\maketitle

\IEEEdisplaynontitleabstractindextext

%
\IEEEpeerreviewmaketitle

\IEEEraisesectionheading{\section{Introduction}\label{intro}}

%
%
%
%
\IEEEPARstart{T}{he} research of micro-expression recognition (MER) aims to enable computers to accurately recognize micro-expressions (MEs) from facial video clips~\cite{oh2018survey,ben2021video,li2022deep}. Unlike ordinary facial expressions, MEs are subtle, rapid, and repressed, often occurring when individuals try to conceal their true emotions. Therefore, MER has significant value in various applications, such as lie detection~\cite{ekman1991can}, and has received extensive attention from researchers in the fields of affective computing, computer vision, and psychology in recent years. However, current MER methods still perform poorly and fail to meet practical application requirements. This is mainly due to the subtle nature of MEs, which correspond to low-intensity facial motion variations~\cite{ekman1969nonverbal}. As a result, any frame in ME video clips appears very similar to its adjacent frames. Thus, deep neural networks, which exhibit expertise in other video-based vision tasks, may fail to be sensitive enough to the minor motion variations associated with MEs, resulting in their inability to learn ME-discriminative features from facial video clips.

To overcome the challenges posed by low-intensity MEs, researchers have proposed several beneficial ideas to enhance the learning of discriminative features for MER. One notable idea can be summarized as "less is more"~\cite{liong2018less}, inspired by Ekman's finding that "a snapshot taken at the peak of an expression can easily convey emotional information"~\cite{ekman1993facial}. Based on this viewpoint, instead of using all frames, we can extract a reduced-size sequence that retains vital and necessary information from the original ME video clip to represent MEs for MER, disregarding the abundance of similar frames caused by low-intensity MEs. For example, Li et al.~\cite{li2018can,li2020joint} attempted to spot the apex frame in ME samples for MER. Their works demonstrate that the deep features learned from apex frames outperform those learned from complete ME sequences. Additionally, the significant works of Liong et al.~\cite{liong2018less,gan2019off} introduce a novel reduced-size structure called the onset-apex structure for representing MEs. In this structure, the onset frame is included with the apex frame to calculate their optical flow representation, facilitating subsequent learning of ME-discriminative features. Compared to using only the apex frame, this structure considers the dynamic information of MEs, thereby learning more discriminative features for MER. Many existing works~\cite{xu2017microexpression,liu2018sparse,verma2019learnet} have demonstrated effectiveness and superior performance of this structure in addressing MER challenges. Recently, Sun et al.~\cite{sun2020dynamic} introduces another reduced-size structure: the onset-apex-offset structure. Compared to the onset-apex structure, this structure additionally incorporates temporal information between the apex and offset frames into the representation of MEs, surpassing the performance of ultilizing all frames for MER tasks.

Another prominent idea is "magnifying facial motions", which involves explicitly magnifying the original ME video clip using video motion magnification techniques. This approach provides a direct solution to overcome the low-intensity interference of MEs. By magnifying the facial movements in MEs, the magnified facial motions become more visibly intense compared to the original movements. This enhancement makes it easier for deep neural networks to learn ME-discriminative features. Representative techniques applied in MER include Eulerian Video Magnification (EVM)~\cite{wu2012eulerian} and its variants, such as Global Lagrangian Motion Magnification (GLMM)~\cite{le2018micro} and Learning-based Video Motion Magnification (LVMM)~\cite{oh2018learning}. Experimental results in many existing works~\cite{zarezadeh2016micro,le2016eulerian,le2018micro,li2018can,lei2020novel,zhi2022micro,ruan2022mimicking} have demonstrated that handcrafted or learned features of magnified ME samples exhibit better discriminability than the original samples. In addition, inspired by the effectiveness of EVM and its variants, some recent deep learning-based MER methods~\cite{lei2020novel,wei2022novel} also incorporate magnification modules into the feature learning process, enabling the learning of more discriminative ME features.

However, it is important to note that the two aforementioned mainstream ideas, "less is more" and "magnifying facial motions", are not always perfect in facilitating the learning of discriminative features for MER. The main reasons can be summarized as follows:

(1) Regarding the "less is more" approach, it has been observed that reduced-size structures, such as onset-apex and onset-apex-offset structures, may still be susceptible to low-intensity interference in MEs and may not effectively learn more discriminative features for MER. This is because the gap between the frames in these structures remains small and does not vary significantly compared to the original ME video clips. Therefore, most existing MER methods based on the "less is more" idea typically incorporate the magnification of facial motions~\cite{li2018can,li2020joint,wei2022novelb}. Additionally, these MER methods rely heavily on accurate apex frame spotting, which limits their flexibility. It is known that accurate apex frame spotting remains a challenging task in ME analysis research and has not yet been sufficiently resolved. Many previous works~\cite{li2020joint,song2019recognizing} have shown that directly using ground truth apex frames for MER methods consistently outperforms using the apex frames spotted by the MER methods themselves.

(2) As for the "magnifying facial motions" approach, determining the appropriate magnification factor for different ME samples often poses an unavoidable issue for corresponding MER methods. Since the intensity of the original ME samples is not visible to us, an unsuitable predetermined magnification factor will inevitably result in excessive magnification and introduce additional noise, such as distortion in some magnified ME video clips~\cite{li2022deep,wei2022novelb}. This diminishes the ability to design or learn discriminative features for MER. Therefore, further research is required to develop more flexible and adaptive magnification methods that can avoid amplification noise in MER methods based on this idea.

\begin{figure}[t!]
\centering
\includegraphics[width=\columnwidth]{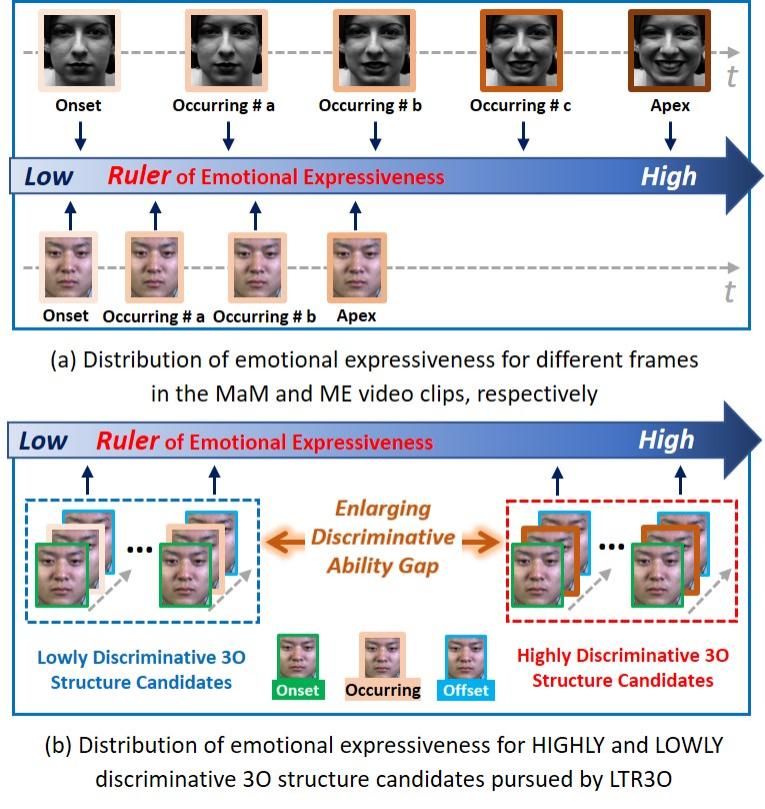}
\caption{Motivation and Basic Idea of the Proposed LTR3O for MER: (a) Comparative Analysis of Emotional Expresiveness Between MaEs and MEs in CK+~\cite{lucey2010extended} and CASME~II~\cite{yan2014casme} Databases. Frames are Randomly Extracted From MaE and ME Video Clips, From Onset to Apex, with a Fixed Interval. Emotional Expressiveness is Manually Compared Based on Expression Discriminative Ability. (b) Illustration of the Proposed LTR3O Idea: Measuring and Calibrating Emotional Expressiveness of 3O Structure Candidates to Align with MaEs Distribution.}
\label{fig:basic_idea}
\end{figure}

\begin{figure*}[t!]
\centering
\includegraphics[width=\textwidth]{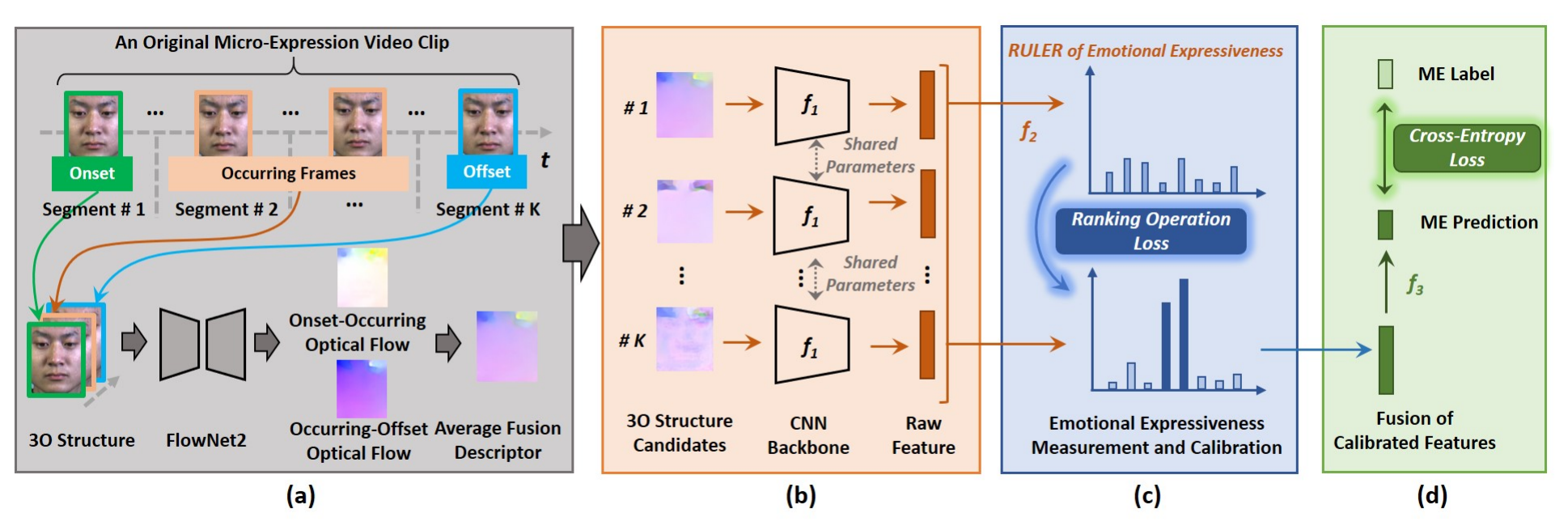}
\caption{Overall Structure of the Proposed LTR3O for MER: (a) Generation of Onset-Occurring-Offset (3O) Structure Candidates and Extraction of Optical Flow Descriptors. (b), (c), and (d) Depict the Three Major Modules in LTR3O: Raw Feature Extraction, Emotional Expressiveness Measurement and Calibration, and ME Prediction with Calibrated 3O Structure Candidates.}
\label{fig_overallstructure}
\end{figure*}

In this paper, we propose a flexible and reliable method called learning to rank onset-occurring-offset representations (LTR3O) for MER. Our method draws on strengths of two constructive ideas mentioned above, while overcoming their limitations. Fig.~\ref{fig:basic_idea} depicts the basic idea behined LTR3O. As shown in Fig.~\ref{fig:basic_idea}(a), unlike MEs, emotional expressiveness of frames in the macro-expression (MaE) increases significantly along the time axis, peaking at apex frame and its neighboring ones. As a result, the MaE can be easily recognized from these salient frames. On the other hand, it can be seen that the ME exhibit a similar trend. However, due to its low-intensity characteristic, the emotional expressiveness of apex and its neighboring frames is insufficient. This means that there is a small emotional expressiveness gap between these frames and the remaining frames in MEs, which differs from the distribution from MaMs. This observation motivates us to calibrate the emotional expressiveness distribution of ME frames to enlarge this gap, making them resemble MaEs and be easily recognized.

To achieve this goal, we raise the idea shown in Fig.~\ref{fig:basic_idea}(b) to guide the design of LTR3O for MER. Specifically, we first present an onset-occurring-offset (3O) structure, consisting of three key frames: onset, occurring, and offset, to represent MEs. Unlike apex frame structure, this reduced-size structure preserves dynamic information in MEs. Moreover, the occurring frame in this structure can be flexibly extracted from the original ME video clips, avoiding the apex frame spotting and increasing the method's flexibility. Then, for each ME sample, we generate a set of 3O structure candidates corresponding to different occurring frames and incorporate a ruler module to measure their emotional expressiveness based on their ME-discriminative ability. The emotional expressiveness distribution of these 3O structure candidates is subsequently calibrated by enlarging the gap between highly discriminative candidates and the remaining. This calibration process aims to make the emotional expressiveness distribution of MEs similar to that of MaEs along the time axis, which ensures that deep neural networks can reliably learn ME-discriminative features from 3O structures of MEs. Extensive experiments conducted on three widely-used ME databases have demonstrated the effectiveness of the proposed LTR3O method. Compared with existing MER methods based on the ideas of "less is more" or "magnifying facial motions", LTR3O has three major advantages:
\begin{enumerate}
\item The 3O structure designed for representing MEs helps manage the abundant information caused by the low-intensity of MEs, while preserving dynamic cues like onset-apex and onset-apex-offset structures.
\item LTR3O avoids the need for apex frame spotting, which is a necessary but challenging prior step required by the apex frame involved MER methods. Instead, we define the occurring frame, which can be flexibly extracted from the original ME video clips, to compose the dynamic reduced-size sequence with onset and offset frames. 
\item LTR3O implicitly magnifies the emotional expressiveness of MEs by enforcing that the emotional expressiveness of 3O structure candidates is distributed similarly to that of MaMs with respect to time. This eliminates the need to consider magnification factors and enables the reliable learning of discriminative features for ME recognition, without being affected by low-intensity interference in MEs.
\end{enumerate}

The rest of this paper is organized as follows: Section~\ref{proposedmethod} provides a detailed description of the proposed LTR3O method and explains how it addresses the MER tasks. In Section~\ref{experiments}, extensive experiments on widely-used ME databases are conducted to evaluate the effectiveness of the proposed LTR3O. Finally, we conclude the paper in Section~\ref{conclusion}.

\section{Proposed Method}
\label{proposedmethod}

\subsection{Overall Picture of LTR3O for MER}

In this section, we present the proposed LTR3O method and describe how it can be applied to the task of MER. The overall framework of LTR3O is illustrated in Fig.~\ref{fig_overallstructure}. Given an ME video clip, LTR3O generates a set of 3O structure candidates and calculates their optical flow (OF) descriptors, which are then utilized as the inputs to the LTR3O model. The LTR3O model consists of three major modules: raw feature extraction, emotional expressiveness measurement and calibration, and ME prediction. These modules collaborate to enable LTR3O to effectively learn ME-discriminative features from the generated 3O structure candidates.

\subsection{A Set of 3O Structure Candidates for Representing MEs}

For a given ME video clip, we partition it into $K$ non-overlapping segments of equal length. From each segment, we randomly extract a frame as the occurring frame and combine it with the onset and offset frames of the original ME video clip to create a 3O structure, as shown in Fig.~\ref{fig_overallstructure}~(a). This procedure generates $K$ 3O structure candidates for a single ME video clip. We then utilize the FlowNet2 model~\cite{ilg2017flownet} to estimate the OF descriptors between the onset and occurring frames, as well as between the occurring and offset frames, for these 3O structure candidates. The two OF representations are subsequently fused using an averaging operation, resulting in a 3-channel pseudocolor image, denoted as $\{\mathcal{X}_{i,j}\}_{j=1}^{K}$, where $i$ and $j$ represent the indices of the ME sample and its associated 3O structure candidate, respectively.

\subsection{Raw Feature Extraction for 3O Structure Candidates}

The first major module of LTR3O is the raw feature extraction, which aims to convert the OF representations associated with a set of 3O structure candidates from the sample space to a raw feature space. To achieve this goal, we employ a set of parameter-shared baseline convolutional neural networks (CNNs), such as ResNet~\cite{he2016deep}, with the same number as the 3O structure candidates. These CNNs serve as the backbones of the raw feature extraction module in LTR3O, as shown in Fig.~\ref{fig_overallstructure}~(b). Let $f_1$ be chosen CNN function. With this setup, we can extract the raw features for the $i^{th}$ ME sample from its corresponding OF representations of $K$ 3O structure candidates, which can be formulated as $\mathbf{x}_{i,j} = f_1(\mathcal{X}_{i,j})$, where $j = 1, \cdots, K$.

\subsection{Emotional Expressiveness Measurement and Calibration for 3O Structure Candidates}

As mentioned earlier, deep neural networks still struggle to learn discriminative features from reduced-size structure. Therefore, it becomes necessary to explicitly magnify the emotional expressions in advance, as the reduced-size structures only remove redundant information while the intensity of emotional expressions remains unchanged. However, determining an appropriate magnification factor for existing video motion magnification methods poses a significant challenge. To address this, we design a reliable module in LTR3O, illustrated in Fig.~\ref{fig_overallstructure}(c). This module measures and calibrates the emotional expressiveness of 3O structure candidates based on their discriminative ability in recognizing MEs. Specifically, we design an attention mechanism-based module consisting of a fully connected (FC) layer and a sigmoid activation function. This module serves as a ruler to measure the specific emotional expressiveness scores of 3O structure candidates. Mathematically, the emotional expressiveness score of the $j^{th}$ 3O structure candidate associated with the $i^{th}$ ME sample, denoted as $\alpha_{i,j}$, can be expressed as follows:
\begin{eqnarray}
\alpha_{i,j} = \frac{\sigma(f_2(\mathbf{x}_{i,j}))}{\sum_{j=1}^K \sigma(f_2(\mathbf{x}_{i,j}))},~j=\{1,\cdots,K\},
\end{eqnarray}
where $\sigma(\cdot)$ represents the sigmoid function, and $f_2$ denotes the operation performed by the FC layer, respectively.

It is worth noting that if we directly used the fusion of raw features, i.e., $\sum_{j=1}^{K}\alpha_{i,j}\mathbf{x}_{i,j}$, to learn LTR3O with the guidance of its corresponding ME label, the obtained values of $\alpha_{i,j}$ measuring their contributions in recognizing MEs would be very close due to the similarity in emotional expressiveness among its corresponding 3O structure candidates. To address this issue, we further enable LTR3O to calibrate the distribution of emotional expressiveness for 3O structure candidates so that they are distributed with a larger intensity gap like the MaM sample shown in Fig.~\ref{fig:basic_idea}. This objective can be achieved by minimizing the following ranking operation loss (RO-Loss) function, which enlarges the gap between highly ME-discriminative 3O structure candidates and the remaining ones:
\begin{eqnarray}
\mathcal{L}_{RO} = \max\{0, \delta - (\alpha_i^h - \alpha_i^l)\}.
\label{eqn:nba}
\end{eqnarray}

In Eq.~(\ref{eqn:nba}), $\alpha_i^h = \frac{1}{K_h} \sum_{j=1}^{K_h}\tilde\alpha_{i,j}$ and $\alpha_i^l = \frac{1}{K-K_h} \sum_{j=K_h + 1}^{K}\tilde\alpha_{i,j}$ represent the mean values of emotional expressiveness associated with the highly and lowly ME-discriminative 3O structure candidates, respectively. The parameter $\delta$ determines the margin that separates the two groups. Here, $\{\tilde\alpha_{i,j}\}_{j=1}^{K} = \{\alpha_{i,j}\}_{j=1}^{K}$ satisfying $\tilde\alpha_{i,1}\geq\tilde\alpha_{i,1}\geq\cdots\geq\tilde\alpha_{i,K}$, $K_h = Ceil(\gamma\times K)$ is the number of highly ME-discriminative 3O structure candidates, and $\gamma$ is the highly ME-discriminative 3O structure candidate ratio, which is a value between 0 and 1. It is important to note that the designed RO-loss in Eq.~(\ref{eqn:nba}) is derived from the triplet loss~\cite{weinberger2009distance}, which is commonly used in learning-to-rank (LTR) research~\cite{cakir2019deep,pretet2020learning}. This is one of the main reasons why we refer to the proposed method as LTR3O.

\subsection{ME Prediction with Calibrated 3O Structure Candidates}

The last module in LTR3O is the ME prediction, shown in Fig.~\ref{fig_overallstructure}~(d). In this module, we start by merging the raw features of $K$ 3O structure candidates by multiplying them with their corresponding calibrated emotional expressiveness scores. This can be formulated as follows:
\begin{eqnarray}
\mathbf{x}_i = \sum_{j=1}^{K}\alpha_{i,j}\mathbf{x}_{i,j}.
\end{eqnarray}
We can then predict the ME label of such ME sample with the fused feature through an FC layer and a softmax operation, expressed as:
\begin{eqnarray}
\mathbf{y}_i^{p} = \textup{softmax}(f_3(\mathbf{x}_i)),
\end{eqnarray}
where $\textup{softmax}(\cdot)$ is the softmax function and $f_3$ represents the operation performed by FC layer. To achieve this, we employ the cross-entropy loss (CE-Loss) to establish the relationship between the ME label predicted by LTR3O and the ground truth. The CE-Loss can be written as:
\begin{eqnarray}
\mathcal{L}_{CE} = \mathcal{J}(\mathbf{y}_i^{g},\mathbf{y}_i^{p}),
\label{eqn:nbb}
\end{eqnarray}
where $\mathcal{J}(\cdot)$ is the cross-entropy function, and $\mathbf{y}_i^{g}$ is the one-hot label vector of the $i^{th}$ ME video clip generated according to its corresponding ground truth.

\subsection{Total Loss Function}

Assume that we have $N$ ME samples for training the LTR3O model. To learn the optimal model parameters of LTR3O, we minimize the following objective function, which is a combination of the CE loss in Eq.~(\ref{eqn:nbb}) and the RO loss in Eq.~(\ref{eqn:nba}), summed over $N$ ME training samples:
\begin{eqnarray}
\min_{\mathcal{W}_{f_1},\mathcal{W}_{f_2},\mathcal{W}_{f_3}} \sum_{i=1}^{N}[\mathcal{J}(\mathbf{y}_i^{g},\mathbf{y}_i^{p}) + \lambda \max\{0,\delta - (\alpha_i^h - \alpha_i^l)\}],
\label{eqn:totalloss}
\end{eqnarray}
where $\lambda$ is a trade-off parameter that controls the balance between the CE and RO losses, and $\mathcal{W}_{f_1}$, $\mathcal{W}_{f_2}$, $\mathcal{W}_{f_3}$ are the parameters of the networks, $f_1$, $f_2$, and $f_3$, used in LTR3O, respectively.

\section{Experiments}
\label{experiments}

\subsection{ME Databases and Experimental Protocol}
\label{databaseandprotocol}
To evaluate the proposed LTR3O method, we carry out extensive experiments using three widely-used ME databases: CASME~II~\cite{yan2014casme}, SMIC~\cite{li2013spontaneous}, and SAMM~\cite{davison2016samm}. CASME~II consists of 247 ME video clips from 26 subjects, recorded by a high-speed camera. Each ME sample is labeled as one of five ME categories: \textit{Disgust} (64 samples), \textit{Happy} (32 samples), \textit{Repressed} (27 samples), \textit{Surprise} (25 samples), and \textit{Others} (99 samples). SMIC has three subsets: HS, VIS, and NIR, recorded by different cameras (high-speed, visual, and near-infrared, respectively). In our experiments, we only utilized the HS subset, which includes 16 subjects and 164 ME samples. Each SMIC sample is labeled as one of three ME categories: \textit{Negative} (70 samples), \textit{Positive} (51 samples), and \textit{Surprise} (43 samples). SAMM, also recorded using a high-speed camera, contains 159 ME image sequences from 32 subjects, belonging to eight ME categories. For our MER experiments, we selected the samples of \textit{Angry} (57 samples), \textit{Contempt} (12 samples), \textit{Happy} (26 samples), \textit{Surprise} (15 samples), and \textit{Others} (26 samples) associated with 27 subjects from SAMM, following the experiment setting of previous works~\cite{khor2019dual, song2019recognizing, li2020joint, lei2020novel, zhi2022micro, ruan2022mimicking}.

We employ the leave-one-subject-out (LOSO) protocol for our MER experiments on all the ME databases. Under this protocol, we conduct $S$ folds of experiments, where $S$ represents the number of subjects in the ME database. In each fold, the ME video clips of a specific subject are used as the testing samples, while the remaining subjects' samples are used for training. The performance metrics chosen in our experiments are \textit{Accuracy} and \textit{F1-Score}, which are calculated as follows: \textit{Accuracy} $= \frac{\sum_{i=1}^{S}T_i}{\sum_{i=1}^{S}N_i}\times 100$, where $T_i$ is the number of the correct predictions for the $i^{th}$ subject, and $N_i$ is the total number of this subject's speech samples, and \textit{F1-Score} $= \frac{1}{C}\sum_{i=1}^{C}\frac{2p_i\times r_i}{p_i+r_i}$, where $p_i$ and $r_i$ represent the precision and recall of the $i^{th}$ ME class, respectively, and $C$ is the total number of ME classes.


\subsection{Comparison Methods and Implementation Detail}
\label{sec:nb2}

In order to evaluate the performance of the proposed LTR3O method in addressing the challenge of MER, we compare it with recent state-of-the-art MER methods, including:
\begin{enumerate}
\item \textit{MER Methods without Using the Ground Truth Apex Frame}: DSSN~\cite{khor2019dual}, TSCNN~\cite{song2019recognizing}, LGCconD~\cite{li2020joint}, SLSTT-LSTM~\cite{zhang2022short}, STLBP-IP + KGSL~\cite{zong2018learning}, OFF-ApexNet~\cite{gan2019off}, STRCN-G~\cite{xia2019spatiotemporal}, TS-AUCNN~\cite{sun2020dynamic}, and KTGSL~\cite{wei2022learning}.
\item \textit{MER Methods Directly Using the Ground Truth Apex Frame}: DSSN~\cite{khor2019dual}, TSCNN~\cite{song2019recognizing}, LGCcon~\cite{li2020joint}, SLSTT-LSTM~\cite{zhang2022short}, MicroNet~\cite{xia2020learning}, TS-AUCNN~\cite{sun2020dynamic}, Graph-TCN~\cite{lei2020novel}, MiNet~\cite{xia2021micro}, MER-Supcon~\cite{zhi2022micro}, GRAPH-AU~\cite{lei2021micro}, Tiny-I3D~\cite{wang2023temporal}, and MAP~\cite{ruan2022mimicking}.
\end{enumerate}

For our experiments, we use ResNet18~\cite{he2016deep} as the CNN backbone for LTR3O. The face images from the original ME video clips are cropped and resized to $112\times 112$ pixels. The implementation of the LTR3O model is done using the PyTorch 1.9 platform, utilizing an NVIDIA RTX 3090 GPU. During training, we employ the Adam optimizer with a batch size of 64. To augment the training data, random horizontal flipping and random resized cropping are applied to generate diverse ME samples. Additionally, we utilize a cosine annealing scheduler to dynamically update the learning rate, which had an initial value of $1e^{-4}$. LTR3O has four important hyperparameters: $K$ (segment number), $\delta$ (margin value), $\gamma$ (highly ME-discriminative 3O structure candidate ratio), and $\lambda$ (trade-off parameter). These hyperparameters are set to fixed values of 8, 0.7, 0.1, and 1, respectively, throughout the experiments on all three ME databases. As for the comparison methods mentioned above, we directly obtain the results from their respective works because they adopt the same experimental protocol as ours.

\subsection{Results and Discussions}

The experimental results are presented in Tables~\ref{tab:nb1} and~\ref{tab:nb2}, which compare the performance of MER methods without and with the use of ground truth apex frames from the ME database, respectively. The first four methods in both tables, highlighted with underlines, necessarily rely on the apex frame and hence provide results for their own spotted frames and direct use of ground truth frames, respectively. Several interesting observations can be made from these tables.

\begin{table*}[t]
\centering
\renewcommand{\arraystretch}{1.3}
\caption{Comparison Results in Terms of \textit{Accuracy} / \textit{F1-Score} with Recent State-of-the-art MER Methods without Using the Ground Truth Apex Frame Provided by the Database. The Best Results for Each Database are Highlighted in Bold.}
\begin{tabular}{|l|c|c|c|c|c|c|}
\hline
\textbf{Method} & \textbf{Year} & \textbf{\makecell[c]{Using Apex \\ Frame GT?}} & \textbf{\makecell[c]{Magnifying Facial \\ Motions in MEs?}} & \textbf{CASME~II} & \textbf{SMIC} & \textbf{SAMM}\\ \hline \hline
\underline{DSSN}~\cite{khor2019dual} &2019 & No &No& N/A & 63.41 / 64.62 & N/A  \\
\underline{TSCNN}~\cite{song2019recognizing} & 2019 & No & No & 74.05 / 73.27 & 72.74 / 72.36 & 63.53 / 60.65  \\
\underline{LGCconD}~\cite{li2020joint} & 2020 & No & Yes (EVM) & 65.05 / 64.00 & 63.41 / 62.00 &35.29 / 23.00  \\
\underline{SLSTT-LSTM}~\cite{zhang2022short}&2022 &No & No &N/A &75.00 / 74.00 &N/A  \\\hline
STLBP-IP + KGSL~\cite{zong2018learning} & 2018 & No & No &65.18 / 62.54 &66.46 / 65.77 & N/A  \\
STRCN-G~\cite{xia2019spatiotemporal}&2019 &No & Yes (EVM)&N/A &72.30 / 69.50&N/A  \\
TS-AUCNN~\cite{sun2020dynamic} &2020 &No &No &N/A & 76.06 / 71.00 & N/A \\
KTGSL~\cite{wei2022learning}&2022&No&Yes (EVM)&72.58 / 68.20&75.64 / 69.17 &56.11 / 49.30 \\\hline\hline
\textbf{LTR3O (Ours)} & \textbf{2023} & \textbf{No} & \textbf{No} &\textbf{78.95} / \textbf{76.46} &\textbf{80.49} / \textbf{80.11} & \textbf{76.47} / \textbf{70.22}  \\\hline
\end{tabular}
\label{tab:nb1}
\end{table*}

\begin{table*}[t]
\centering
\renewcommand{\arraystretch}{1.3}
\caption{Comparison Results in Yerms of \textit{Accuracy} / \textit{F1-Score} with Recent State-of-the-art MER Methods Directly Using the Ground Truth Apex Frame Provided by the Database. The Best Results for Each Database are Highlighted in Bold.}
\begin{tabular}{|l|c|c|c|c|c|c|}
\hline
\textbf{Method} & \textbf{Year} & \textbf{\makecell[c]{Using Apex \\ Frame GT?}} & \textbf{\makecell[c]{Magnifying Facial \\ Motions in MEs?}} & \textbf{CASME~II} & \textbf{SMIC} & \textbf{SAMM}\\ \hline \hline
\underline{DSSN}~\cite{khor2019dual} &2019 & Yes &No &70.78 / 72.97 & N/A & 57.35 / 46.44  \\
\underline{TSCNN}~\cite{song2019recognizing} &2019 & Yes & No & 80.97 / 80.70 & N/A & 71.76 / 69.42  \\
\underline{LGCcon}~\cite{li2020joint} &2020 & Yes & Yes (EVM) &65.02 / 64.00 &N/A & 40.90 / 34.00  \\
\underline{SLSTT-LSTM}~\cite{zhang2022short} &2022&Yes &No &75.81 / 75.30 &N/A &72.39 / 64.00  \\\hline
MicroNet~\cite{xia2020learning} & 2020 & Yes & No &75.60 / 70.10 & 76.80 / 74.40 & 74.10 / 73.60  \\
TS-AUCNN~\cite{sun2020dynamic} &2020 &Yes & No & 72.61 / 67.00 & N/A & N/A  \\
Graph-TCN~\cite{lei2020novel} &2020 &Yes &Yes (LVMM)&73.98 / 72.46 & N/A &75.00 / 69.85  \\
MiNet~\cite{xia2021micro}& 2021 &Yes & No &79.90 / 75.90 & 78.60 / 77.80 &76.70 / \textbf{76.40} \\
MER-Supcon~\cite{zhi2022micro} &2022&Yes&Yes (LVMM) &73.58 / 72.86&N/A& 67.65 / 62.51  \\
MAP~\cite{ruan2022mimicking}&2022&Yes&Yes (LVMM) &\textbf{83.30} / \textbf{82.70} &N/A &\textbf{79.40} / 75.80   \\\hline\hline
\textbf{LTR3O (Ours)} & \textbf{2023} & \textbf{No} & \textbf{No} & 78.95 / 76.46 &\textbf{80.49} / \textbf{80.11} & 76.47 / 70.22  \\\hline
\end{tabular}
\label{tab:nb2}
\end{table*}

Firstly, Table~\ref{tab:nb1} clearly demonstrates that our LTR3O method achieves the best results in terms of \textit{Accuracy} and \textit{F1-Score} on all three ME databases among methods that do not use ground truth apex frames. Even compared to methods that utilize EVM to enhance original ME video clips in advance, LTR3O consistently outperforms them on all databases with remarkable improvements. These findings highlight LTR3O as a flexible and high-performing MER method that does not rely on apex frame spotting or pre-processing facial motion magnification.

Secondly, comparing the results for the first four methods in both tables, it is evident that using spotted apex frames leads to a noticeable decrease in performance compared to using ground truth frames directly. This emphasizes the importance of accurately spotting apex frame for apex frame involved deep learning methods to learn discriminative features for MER. Moreover, it also suggests that these methods have limited flexibility and reliability, as accurately spotting the apex frame remains a challenging task that has not been fully addressed.

Lastly, in Table~\ref{tab:nb2}, most methods that directly use ground truth apex frames, including LGCcon, Graph-TCN, MER-Supcon, and MAP, employ video motion magnification techniques to enhance facial motions in MEs, resulting in promising MER performance. However, our LTR3O method, which only requires the original ME video clips, performs better than three of these methods in terms of both \textit{Accuracy} and \textit{F1-Score}. Although MAP outperforms LTR3O on the CASME II and SAMM databases, it is important to note that LTR3O still demonstrates strong competitiveness against MAP. Additionally, it should be mentioned that MAP relies on accurate apex frame information, and its performance may suffer if the apex frame was not accurately spotted by existing automatic spotting methods. Considering these factors, LTR3O proves to be a more flexible MER method with satisfactory performance.

\begin{table*}[t!]
	\centering
	\renewcommand{\arraystretch}{1.3}
	\caption{Comparison Results in Terms of \textit{UF1} / \textit{UAR} with Recent State-of-the-art MER Methods under the CDE Protocol. The Best Results are Highlighted in Bold.}
	\begin{tabular}{|l|c|c|c|ccc|c|}
		\hline
		\textbf{Method} & \textbf{Year} & \textbf{\makecell[c]{Using Apex \\ Frame GT?}} & \textbf{\makecell[c]{Magnifying Facial \\ Motions in MEs?}}& \textbf{CASME~II} & \textbf{SMIC} & \textbf{SAMM}  & \textbf{Composite} \\ \hline \hline
		OFF-ApexNet~\cite{gan2019off}&2019&Yes&No&87.64 / 86.81& 68.17 /66.95 &54.09 / 53.92&71.96 / 70.96\\
		MicroNet~\cite{xia2020learning} & 2020 & Yes & No &87.00 / 87.20&86.40 / 86.10&82.50 / 81.90&86.40 / 85.70  \\
		GRAPH-AU~\cite{lei2021micro}&2021&Yes &Yes (LVMM) &87.98 / 87.10& 71.92 / 72.15 &77.51 / 78.90 &79.14 / 79.33\\
		MiNet (CK+)~\cite{xia2021micro}& 2021 &Yes & No  & 88.10 / 88.10 &\textbf{87.30} / \textbf{86.70}&\textbf{89.60} / \textbf{88.40} & 88.30 / \textbf{87.60}\\
		MiNet (MMI)~\cite{xia2021micro}& 2021 &Yes & No  & N/A & N/A & N/A & 86.30 / 86.00\\
		MERSiamC3D~\cite{zhao2021two}&2021&Yes&No&88.18 / 87.63& 73.56 / 75.98 &74.75 / 72.80 &80.68 / 79.86\\
		SLSTT-LSTM~\cite{zhang2022short} &2022&Yes &No &90.10 / 88.50& 74.00 / 72.00 &71.50 / 64.30 &81.60 / 79.00  \\
		Tiny-I3D~\cite{wang2023temporal}&2023&Yes&No&\textbf{93.70} / \textbf{92.71}& 77.39 / 75.84&79.19 / 74.04  &83.69 / 80.92 \\ 
\hline\hline
\textbf{LTR3O (Ours)} & \textbf{2023} & \textbf{No} & \textbf{No}  &93.30 / 91.78&87.00 / 86.60 & 85.13 / 80.47 & \textbf{89.13} / 87.44 \\\hline
	\end{tabular}
	\label{tab:nb3}
\end{table*}

\begin{figure*}[!t]
\centering
\includegraphics[width=0.8\textwidth]{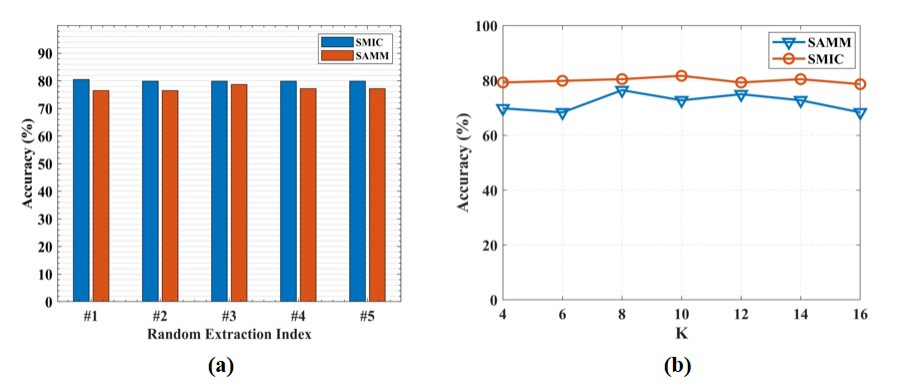}
\caption{Results in Terms of \textit{Accuracy} for Investigating the Reliability of LTR3O: (a) Segment Number $K$ is Fixed at 8 and the Occurring Frame is Randomly Extracted From Each Segment Five Times. (b) Values of Segment Number $K$ Varies From 4 to 16 with an Interval of 2.
}
\label{fig_reliability}
\end{figure*}

\subsection{Comparison Results for the MEGC Challenge}

We also evaluate our LTR3O method for addressing the MER tasks under the procotol of composite database evaluation (CDE) adpoated in the MEGC challenge~\cite{see2019megc}. In contrast to LOSO, CDE provides a more realistic evaluation scenario, in which subjects come from diverse backgrounds, such as ethnicity, for MER methods. In CDE, the CASME~II, SMIC, and SAMM databases are combined into a single composite database based on the shared ME classes: \textit{Positive}, \textit{Negative}, and \textit{Surprise}, including 68 subjects (16 from SMIC, 24 from CASME II, and 28 from SAMM). Then, LOSO is applied to determine the training and testing spits in the experiments on this composite database. The official performance metrics for this challenge are \textit{unweighted F1-score (UF1)} and \textit{unweighted average recall (UAR)}, where \textit{UF1} is equivalent to the \textit{F1-Score} defined in Section~\ref{databaseandprotocol}, while \textit{UAR} is defined as the average accuracy across the ME classes. The comparison methods include recent state-of-the-art MER methods: OFF-ApexNet~\cite{gan2019off}, MicroNet~\cite{xia2020learning}, GRAPH-AU~\cite{lei2021micro}, MiNet~\cite{xia2021micro}, MERSiamC3D~\cite{zhao2021two}, SLSTT-LSTM~\cite{zhang2022short}, and Tiny-I3D~\cite{wang2023temporal}.

Table~\ref{tab:nb3} presents the experimental results. It is evident from the table that our LTR3O method achieves the best results in terms of \textit{UF1} among all the comparison methods, demonstrating the effectiveness and superior performance in coping with the MER tasks under the CDE protocol. Although our LTR3O method does not surpass MiNet in terms of UAR, the discrepancy between their results is minimal. Moreover, according to the work of~\cite{xia2021micro}, it is known that MiNet leverages the knowledge of MaEs to guide the networks to be more aware of the facial motion variation. Therefore, it requires MaE samples to jointly learn features for recognizing MEs alongside ME samples. It is clear from Table~\ref{tab:nb3} that the performance of MiNet varies with respect to the choice of the macro-expression database, decreasing from 88.30/87.60 to 86.30/86.00 when the MaE database CK+~\cite{lucey2010extended} is replaced by MMI~\cite{valstar2010induced}. Additionally, it should be noted that unlike all comparison methods including MiNet, our LTR3O method does not directly utilize ground truth apex frames provided by these ME databases, further demonstrating the flexibility and more comprehensive performance of the proposed LTR3O method.

\begin{figure*}[!t]
\centering
\includegraphics[width=\textwidth]{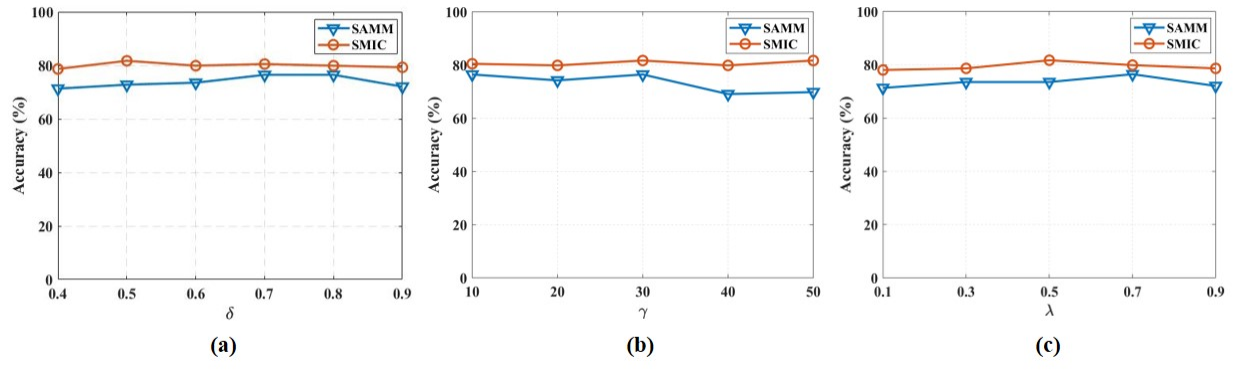}
\caption{Results in Terms of \textit{Accuracy} for the Experiments of Going Deeper Into the Emotional Expressiveness Measurement and Calibration Module in LTR3O. From Left to Right, the Results Correspond to How the Performance of LTR3O Varies with Respect to the Changes of (a) $\delta$ (Margin Value), (b) $\gamma$ (Highly ME-Discriminative 3O Structure Candidate Ratio), and (c) $\lambda$ (Trade-off Parameter Balancing the Module Loss and Cross-Entropy Loss), Respectively. 
}
\label{fig:goingdeeper}
\end{figure*}

\begin{table*}[!t]
\centering
\renewcommand{\arraystretch}{1.3}
\caption{Comparison Results in Terms of \textit{Accuracy} / \textit{F1-Score} Between 3O and Other Reduced-size Structures Used for MER. The Best Results Among All the Types of Structures are Highlighted in Bold and Among Each Type are Highlighted with the Underline.}
\begin{tabular}{|c|c|c|c|c|c|c|}
\hline
\multicolumn{2}{|c|}{\textbf{Reduced-Size Structure}} & \textbf{Baseline Model} & \textbf{\makecell[c]{Using\\Measurement?}} & \textbf{\makecell[c]{Using\\Calibration?}} & \textbf{SMIC} & \textbf{SAMM} \\ \hline \hline
\multirow{7}{*}{\makecell[c]{Single-Frame\\Structure\\(Static)}} & Apex Frame & ResNet18 & - & - & - & \underline{72.79} / \underline{65.72} \\
& 1O Structure\#1 & ResNet18 & - & - &71.95 / 71.09 & 72.06 / 65.40 \\
& 1O Structure\#3 & ResNet18 & - & - & \underline{72.56} / 70.89 &72.06 / 65.14 \\
& 1O Structure\#5 & ResNet18 & - & - &71.34 / 69.57 &71.32 / 59.88 \\
& 1O Structure\#7 & ResNet18 & - & - &68.90 / 68.77 &69.85 / 58.14 \\
& \{1O Candidates\} & LTR3O (ResNet18) & Yes & No & 64.63 / 64.50 &68.38 / 58.48 \\
& \{1O Candidates\} & LTR3O (ResNet18) & Yes & Yes & 71.95 / \underline{71.40} & \underline{72.79} / 61.20 \\\hline\hline
\multirow{7}{*}{\makecell[c]{Two-Frame\\Structure\\(Dynamic)}} & Onset-Apex & ResNet18 & - & - & - & 74.26 / 69.70\\
& 2O Structure\#1 & ResNet18 & - & - &74.39 / 74.10 &72.79 / 65.43 \\
& 2O Structure\#3 & ResNet18 & - & - &77.44 / 78.08 &72.06 / \underline{69.81} \\
& 2O Structure\#5 & ResNet18 & - & - &74.39 / 74.83 & 74.26 / 66.08\\
& 2O Structure\#7 & ResNet18 & - & - &73.78 / 73.65 &66.18 / 60.06 \\
& \{2O Candidates\} & LTR3O (ResNet18) & Yes & No & 76.22 / 76.85 & 73.53 / 68.90 \\
& \{2O Candidates\} & LTR3O (ResNet18) & Yes & Yes & \underline{78.66} / \underline{78.98} & \underline{75.00} / 69.36 \\\hline\hline
\multirow{7}{*}{\makecell[c]{Three-Frame\\Structure\\(Dynamic)}} & Onset-Apex-Offset & ResNet18 & - & - & - &73.53 / 68.06 \\
& 3O Structure\#1 & ResNet18 & - & - & 74.39 / 73.76&72.79 / 65.38 \\
& 3O Structure\#3 & ResNet18 & - & - &77.44 / 76.83 & 74.26 / 68.06\\
& 3O Structure\#5 & ResNet18 & - & - &75.00 / 75.64 &72.06 / 63.68 \\
& 3O Structure\#7 & ResNet18 & - & - &75.00 / 74.32 &73.53 / 67.71 \\
& \{3O Candidates\} & LTR3O (ResNet18) & Yes & No & 77.44 / 77.57 & 71.32 / 63.20\\
& \{3O Candidates\} & LTR3O (ResNet18) & Yes & Yes & \underline{\textbf{80.49}} / \underline{\textbf{80.11}} & \underline{\textbf{76.47}} / \underline{\textbf{70.22}}\\\hline
\end{tabular}
\label{tab:nb5}
\end{table*}

\subsection{Delving Deeper into LTR3O for MER}
\subsubsection{Investigating the Reliability of LTR3O}

As demonstrated in the previous experiments, one significant advantage of LTR3O is its flexibility, as it does not heavily rely on apex frame, whose spotting task currently remains challenging. Instead, LTR3O utilizes a 3O structure that efficiently represents MEs by extracting only the occurring frames, without the need for a specific frame spotting method. Therefore, it is crucial to examine the reliability of the randomly extracted occurring frames employed in LTR3O. In order to achieve this, we propose to conduct experiments to investigate how the performance of LTR3O is affected by two factors associated with occurring frames: (1) variations in the number of segments, denoted as $K$, set for randomly extracting occurring frames, and (2) stability of results corresponding to multiple times of random occurring frame extraction within a fixed number of segments.

The experimental results are depicted in Fig.~\ref{fig_reliability}, where (a) and (b) correspond to the results associated with the two aforementioned factors, respectively. In (a), we fixed the segment number $K$ at 8 and randomly generated the occurring frame index for each segment four additional times. Using the extracted occurring frames, we conducted experiments of LTR3O on the SMIC and SAMM databases. For (b), we utilized LTR3O to conduct experiments on both ME databases, varying the values of segment number $K$ from 4 to 16. As illustrated in Fig.~\ref{fig_reliability}, the performance of LTR3O shows minimal variation among the five cases of random occurring frame extraction and remains consistent despite changes in the segment number. These findings provide evidence supporting the reliability of the proposed LTR3O method.


\subsubsection{Evaluation of the Emotional Expressiveness Measurement and Calibration Module in LTR3O}

The reliability of the proposed LTR3O, as demonstrated in the previous experiments, is largely due to the well-designed emotional expressiveness measurement and calibration module in LTR3O. This module has three important hyperparameters that need to be set beforehand for LTR3O to function: $\delta$ (margin value), $\gamma$ (highly ME-discriminative 3O structure candidate ratio), and $\lambda$ (trade-off parameter balancing the module loss and cross-entropy loss). Therefore, we conduct additional experiments on SMIC and SAMM databases to investigate how the performance of LTR3O for MER is affected when adjusting these three hyperparameters. We changed one hyperparameter at a time while fixing the others, with $\delta$ values ranging from 0.4 to 0.9 with an interval of 0.1, $\gamma$ values ranging from 10\% to 50\% with an interval of 10\%, and $\lambda$ values ranging from 0.1 to 0.9 with an interval of 0.1. The experimental results are shown in Fig.~\ref{fig:goingdeeper}. From these results, we observed that while the performance of LTR3O varies with changes in these three hyperparameters, it is overall less sensitive to their choice. Therefore, it is not necessary to meticulously select these three hyperparameters to enable LTR3O to flexibly and reliably learn ME-discriminative features from 3O structure candidates.

\subsubsection{Evaluation of the 3O Structure in LTR3O}

In LTR3O, the occurring frames can effectively facilitate the flexible and reliable learning of ME-discriminative features, as their corresponding 3O structure provides an efficient representation of ME. This raises an interesting question: Is the 3O structure truly a satisfactory reduced-size structure for representing ME? To address this issue, we aim to compare our 3O structure with other widely-used reduced-size structures and conduct comparative experiments on the SMIC and SAMM databases. Specifically, we include the following three types of reduced-size structures in the comparison:
\begin{enumerate}
\item \textit{Single-Frame Structure} (Static): Apex Frame, 1O Structure\#1, \#3, \#5, \#7 (the 1$^{st}$, 2$^{nd}$, 3$^{th}$, and 7$^{th}$ occurring frames randomly extracted from $K=8$ segments for LTR3O set in Section~\ref{sec:nb2}), and \{1O Candidates\} (a set containing all eight occurring frames).
\item \textit{Two-Frame Structure} (Dynamic): Onset-Apex (the optical flow representation calculated between the onset and apex frames), 2O Structure\#1, \#3, \#5, \#7 (the optical flow representations calculated between the onset and the occurring frame randomly extracted from the 1$^{st}$, 2$^{nd}$, 3$^{th}$, and 7$^{th}$ segment, respectively, for LTR3O set in Section~\ref{sec:nb2}), and \{2O Candidates\} (a set of optical flow representations calculated between the onset and one of all eight occurring frames, respectively).
\item \textit{Three-Frame Structure} (Dynamic): Onset-Apex-Offset, 3O Structure\#1, \#3, \#5, \#7 (the optical flow representations calculated among the onset, offset, and occurring frames randomly extracted from the 1$^{st}$, 2$^{nd}$, 3$^{th}$, and 7$^{th}$ segment, respectively, for LTR3O set in Section~\ref{sec:nb2}), and \{3O Candidates\} (a set of optical flow representations calculated among the onset, offset, and one of all eight occurring frames, respectively).
\end{enumerate}

We also employ ResNet18 as the baseline model to learn features from these reduced-size structures. It is worth noting that for \{1O candidates\}, \{2O candidates\}, and \{3O candidates\}, we fuse the features by using LTR3O with ResNet18 as the CNN backbone. The experimental results are shown in Table~\ref{tab:nb5}. Several interesting observations and conclusions can be drawn from Table~\ref{tab:nb5}:

Firstly, in both the comparison between 1O/2O/3O structures and the comparison between apex/onset-apex/onset-apex-offset structures, features learned from 2O/3O/onset-apex-offset structures (\textit{Dynamic}) outperform those from 1O/apex structures (\textit{Static}). This suggests that considering the dynamic information in ME is indeed necessary for representing MEs. Additionally, it can be observed that our designed 3O and onset-apex-offset structure perform better than the 2O and onset-apex structures, even though they both consider the dynamic information in MEs. This might be because that the 3O and onset-apex-offset structures consider the complete dynamic information from start to occurring/apex to end, while the 2O and onset-apex structure only involve incomplete and partial dynamic information. In summary, these observations indicate that compared to the static information in MEs (single-frame structures), incorporating the complete dynamic information characterizing facial movements in MEs can facilitate the learning of ME-discriminative features.

Secondly, it is evident from the comparison between all three types of reduced-size structures (single-frame/two-frame/three-frame) that the learned features involving occurring frames generally perform worse than those involving apex frames, indicating that apex frames inherently contain more informative and essential ME-related discriminative cues. Hence, our comparative results serve as additional empirical evidence supporting the feasibility of MER approaches based on apex frames. However, by introducing more 1O/2O/3O structure candidates, i.e., \{1O candidates\}, \{2O candidates\}, and \{3O candidates\}, and fusing them with LTR3O, remarkble performance improvements are observed for all three reduced-size structures compared to using a single candidate alone, as well as the apex frame. Particularly, the proposed 3O structure (LTR3O) achieves the best results. This further demonstrates the effectiveness and reliablity of LTR3O for MER, benefiting from the flexible 3O structure for representing ME and its emotional expressiveness measurement and calibration module.

\section{Conclusion}
\label{conclusion}

In this paper, we have proposed a novel method called LTR3O to address the issue of MER by mitigating the interference caused by the low-intensity facial motions in MEs. Our approach combines the strengths of existing concepts such as "less is more" and "facial motion magnification", while overcoming their limitations, i.e., the need for accurate spotting of the apex frame in advance and the potential introduction of magnification noise. To achieve this, we introduce a new frame called the occurring frame, which can be flexibly extracted from the original ME video clip. This frame, along with the onset and offset frames, forms a reduced-size structure called 3O for effectively representing MEs and facilitating the learning of ME-discriminative features. Additionally, we design an emotional expressiveness measurement and calibration module, allowing LTR3O to reliably learn ME-discriminative features from a set of 3O structure candidates in MEs by leveraging their distribution similar to MaMs with respect to time. Extensive experiments conducted on three publicly available ME databases demonstrate that our proposed LTR3O method outperforms recent state-of-the-art MER methods. Importantly, LTR3O offers considerable flexibility, as it does not rely on accurate apex frame spotting or video motion magnification techniques. It even achieves superior performance compared to many MER methods that utilize ground truth apex information or employ EVM to magnify facial motions in MEs beforehand.

\ifCLASSOPTIONcompsoc
%
%

\ifCLASSOPTIONcaptionsoff
\newpage
\fi



\bibliographystyle{IEEEtran}
\bibliography{TAFFC2023}

\begin{thebibliography}{10}
\providecommand{\url}[1]{#1}
\csname url@samestyle\endcsname
\providecommand{\newblock}{\relax}
\providecommand{\bibinfo}[2]{#2}
\providecommand{\BIBentrySTDinterwordspacing}{\spaceskip=0pt\relax}
\providecommand{\BIBentryALTinterwordstretchfactor}{4}
\providecommand{\BIBentryALTinterwordspacing}{\spaceskip=\fontdimen2\font plus
\BIBentryALTinterwordstretchfactor\fontdimen3\font minus
  \fontdimen4\font\relax}
\providecommand{\BIBforeignlanguage}[2]{{%
\expandafter\ifx\csname l@#1\endcsname\relax
\typeout{** WARNING: IEEEtran.bst: No hyphenation pattern has been}%
\typeout{** loaded for the language `#1'. Using the pattern for}%
\typeout{** the default language instead.}%
\else
\language=\csname l@#1\endcsname
\fi
#2}}
\providecommand{\BIBdecl}{\relax}
\BIBdecl

\bibitem{oh2018survey}
Y.-H. Oh, J.~See, A.~C. Le~Ngo, R.~C.-W. Phan, and V.~M. Baskaran, ``A survey
  of automatic facial micro-expression analysis: databases, methods, and
  challenges,'' \emph{Frontiers in Psychology}, vol.~9, p. 1128, 2018.

\bibitem{ben2021video}
X.~Ben, Y.~Ren, J.~Zhang, S.-J. Wang, K.~Kpalma, W.~Meng, and Y.-J. Liu,
  ``Video-based facial micro-expression analysis: A survey of datasets,
  features and algorithms,'' \emph{IEEE Transactions on Pattern Analysis and
  Machine Intelligence}, 2021.

\bibitem{li2022deep}
Y.~Li, J.~Wei, Y.~Liu, J.~Kauttonen, and G.~Zhao, ``Deep learning for
  micro-expression recognition: A survey,'' \emph{IEEE Transactions on
  Affective Computing}, 2022.

\bibitem{ekman1991can}
P.~Ekman and M.~O'Sullivan, ``Who can catch a liar?'' \emph{American
  psychologist}, vol.~46, no.~9, p. 913, 1991.

\bibitem{ekman1969nonverbal}
P.~Ekman and W.~V. Friesen, ``Nonverbal leakage and clues to deception,''
  \emph{Psychiatry}, vol.~32, no.~1, pp. 88--106, 1969.

\bibitem{liong2018less}
S.-T. Liong, J.~See, K.~Wong, and R.~C.-W. Phan, ``Less is more:
  Micro-expression recognition from video using apex frame,'' \emph{Signal
  Processing: Image Communication}, vol.~62, pp. 82--92, 2018.

\bibitem{ekman1993facial}
P.~Ekman, ``Facial expression and emotion.'' \emph{American psychologist},
  vol.~48, no.~4, p. 384, 1993.

\bibitem{li2018can}
Y.~Li, X.~Huang, and G.~Zhao, ``Can micro-expression be recognized based on
  single apex frame?'' in \emph{ICIP}.\hskip 1em plus 0.5em minus 0.4em\relax
  IEEE, 2018, pp. 3094--3098.

\bibitem{li2020joint}
------, ``Joint local and global information learning with single apex frame
  detection for micro-expression recognition,'' \emph{IEEE Transactions on
  Image Processing}, vol.~30, pp. 249--263, 2020.

\bibitem{gan2019off}
Y.~S. Gan, S.-T. Liong, W.-C. Yau, Y.-C. Huang, and L.-K. Tan, ``Off-apexnet on
  micro-expression recognition system,'' \emph{Signal Processing: Image
  Communication}, vol.~74, pp. 129--139, 2019.

\bibitem{xu2017microexpression}
F.~Xu, J.~Zhang, and J.~Z. Wang, ``Microexpression identification and
  categorization using a facial dynamics map,'' \emph{IEEE Transactions on
  Affective Computing}, vol.~8, no.~2, pp. 254--267, 2017.

\bibitem{liu2018sparse}
Y.-J. Liu, B.-J. Li, and Y.-K. Lai, ``Sparse mdmo: Learning a discriminative
  feature for micro-expression recognition,'' \emph{IEEE Transactions on
  Affective Computing}, vol.~12, no.~1, pp. 254--261, 2018.

\bibitem{verma2019learnet}
M.~Verma, S.~K. Vipparthi, G.~Singh, and S.~Murala, ``Learnet: Dynamic imaging
  network for micro expression recognition,'' \emph{IEEE Transactions on Image
  Processing}, vol.~29, pp. 1618--1627, 2019.

\bibitem{sun2020dynamic}
B.~Sun, S.~Cao, D.~Li, J.~He, and L.~Yu, ``Dynamic micro-expression recognition
  using knowledge distillation,'' \emph{IEEE Transactions on Affective
  Computing}, vol.~13, no.~2, pp. 1037--1043, 2020.

\bibitem{wu2012eulerian}
H.-Y. Wu, M.~Rubinstein, E.~Shih, J.~Guttag, F.~Durand, and W.~Freeman,
  ``Eulerian video magnification for revealing subtle changes in the world,''
  \emph{ACM Transactions on Graphics}, vol.~31, no.~4, pp. 1--8, 2012.

\bibitem{le2018micro}
A.~C. Le~Ngo, A.~Johnston, R.~C.-W. Phan, and J.~See, ``Micro-expression motion
  magnification: Global lagrangian vs. local eulerian approaches,'' in
  \emph{FG}.\hskip 1em plus 0.5em minus 0.4em\relax IEEE, 2018, pp. 650--656.

\bibitem{oh2018learning}
T.-H. Oh, R.~Jaroensri, C.~Kim, M.~Elgharib, F.~Durand, W.~T. Freeman, and
  W.~Matusik, ``Learning-based video motion magnification,'' in \emph{ECCV},
  2018, pp. 633--648.

\bibitem{zarezadeh2016micro}
E.~Zarezadeh and M.~Rezaeian, ``Micro expression recognition using the eulerian
  video magnification method,'' \emph{BRAIN. Broad Research in Artificial
  Intelligence and Neuroscience}, vol.~7, no.~3, pp. 43--54, 2016.

\bibitem{le2016eulerian}
A.~C. Le~Ngo, Y.-H. Oh, R.~C.-W. Phan, and J.~See, ``Eulerian emotion
  magnification for subtle expression recognition,'' in \emph{ICASSP}.\hskip
  1em plus 0.5em minus 0.4em\relax IEEE, 2016, pp. 1243--1247.

\bibitem{lei2020novel}
L.~Lei, J.~Li, T.~Chen, and S.~Li, ``A novel graph-tcn with a graph structured
  representation for micro-expression recognition,'' in \emph{ACM Multimedia},
  2020, pp. 2237--2245.

\bibitem{zhi2022micro}
R.~Zhi, J.~Hu, and F.~Wan, ``Micro-expression recognition with supervised
  contrastive learning,'' \emph{Pattern Recognition Letters}, vol. 163, pp.
  25--31, 2022.

\bibitem{ruan2022mimicking}
B.-K. Ruan, L.~Lo, H.-H. Shuai, and W.-H. Cheng, ``Mimicking the annotation
  process for recognizing the micro expressions,'' in \emph{ACM Multimedia},
  2022, pp. 228--236.

\bibitem{wei2022novel}
M.~Wei, W.~Zheng, X.~Jiang, Y.~Zong, C.~Lu, and J.~Liu, ``A novel
  magnification-robust network with sparse self-attention for micro-expression
  recognition,'' in \emph{ICPR}.\hskip 1em plus 0.5em minus 0.4em\relax IEEE,
  2022, pp. 1120--1126.

\bibitem{wei2022novelb}
M.~Wei, W.~Zheng, Y.~Zong, X.~Jiang, C.~Lu, and J.~Liu, ``A novel
  micro-expression recognition approach using attention-based
  magnification-adaptive networks,'' in \emph{ICASSP}.\hskip 1em plus 0.5em
  minus 0.4em\relax IEEE, 2022, pp. 2420--2424.

\bibitem{song2019recognizing}
B.~Song, K.~Li, Y.~Zong, J.~Zhu, W.~Zheng, J.~Shi, and L.~Zhao, ``Recognizing
  spontaneous micro-expression using a three-stream convolutional neural
  network,'' \emph{IEEE Access}, vol.~7, pp. 184\,537--184\,551, 2019.

\bibitem{lucey2010extended}
P.~Lucey, J.~F. Cohn, T.~Kanade, J.~Saragih, Z.~Ambadar, and I.~Matthews, ``The
  extended cohn-kanade dataset (ck+): A complete dataset for action unit and
  emotion-specified expression,'' in \emph{CVPR Workshops}.\hskip 1em plus
  0.5em minus 0.4em\relax IEEE, 2010, pp. 94--101.

\bibitem{yan2014casme}
W.-J. Yan, X.~Li, S.-J. Wang, G.~Zhao, Y.-J. Liu, Y.-H. Chen, and X.~Fu,
  ``Casme ii: An improved spontaneous micro-expression database and the
  baseline evaluation,'' \emph{PloS one}, vol.~9, no.~1, p. e86041, 2014.

\bibitem{ilg2017flownet}
E.~Ilg, N.~Mayer, T.~Saikia, M.~Keuper, A.~Dosovitskiy, and T.~Brox, ``Flownet
  2.0: Evolution of optical flow estimation with deep networks,'' in
  \emph{Proceedings of the IEEE conference on computer vision and pattern
  recognition}, 2017, pp. 2462--2470.

\bibitem{he2016deep}
K.~He, X.~Zhang, S.~Ren, and J.~Sun, ``Deep residual learning for image
  recognition,'' in \emph{Proceedings of the IEEE conference on computer vision
  and pattern recognition}, 2016, pp. 770--778.

\bibitem{weinberger2009distance}
K.~Q. Weinberger and L.~K. Saul, ``Distance metric learning for large margin
  nearest neighbor classification.'' \emph{Journal of machine learning
  research}, vol.~10, no.~2, 2009.

\bibitem{cakir2019deep}
F.~Cakir, K.~He, X.~Xia, B.~Kulis, and S.~Sclaroff, ``Deep metric learning to
  rank,'' in \emph{Proceedings of the IEEE/CVF conference on computer vision
  and pattern recognition}, 2019, pp. 1861--1870.

\bibitem{pretet2020learning}
L.~Pr{\'e}tet, G.~Richard, and G.~Peeters, ``Learning to rank music tracks
  using triplet loss,'' in \emph{ICASSP 2020-2020 IEEE International Conference
  on Acoustics, Speech and Signal Processing (ICASSP)}.\hskip 1em plus 0.5em
  minus 0.4em\relax IEEE, 2020, pp. 511--515.

\bibitem{li2013spontaneous}
X.~Li, T.~Pfister, X.~Huang, G.~Zhao, and M.~Pietik{\"a}inen, ``A spontaneous
  micro-expression database: Inducement, collection and baseline,'' in
  \emph{FG}.\hskip 1em plus 0.5em minus 0.4em\relax IEEE, 2013, pp. 1--6.

\bibitem{davison2016samm}
A.~K. Davison, C.~Lansley, N.~Costen, K.~Tan, and M.~H. Yap, ``Samm: A
  spontaneous micro-facial movement dataset,'' \emph{IEEE Transactions on
  Affective Computing}, vol.~9, no.~1, pp. 116--129, 2016.

\bibitem{khor2019dual}
H.-Q. Khor, J.~See, S.-T. Liong, R.~C. Phan, and W.~Lin, ``Dual-stream shallow
  networks for facial micro-expression recognition,'' in \emph{ICIP}.\hskip 1em
  plus 0.5em minus 0.4em\relax IEEE, 2019, pp. 36--40.

\bibitem{zhang2022short}
L.~Zhang, X.~Hong, O.~Arandjelovi{\'c}, and G.~Zhao, ``Short and long range
  relation based spatio-temporal transformer for micro-expression
  recognition,'' \emph{IEEE Transactions on Affective Computing}, vol.~13,
  no.~4, pp. 1973--1985, 2022.

\bibitem{zong2018learning}
Y.~Zong, X.~Huang, W.~Zheng, Z.~Cui, and G.~Zhao, ``Learning from hierarchical
  spatiotemporal descriptors for micro-expression recognition,'' \emph{IEEE
  Transactions on Multimedia}, vol.~20, no.~11, pp. 3160--3172, 2018.

\bibitem{xia2019spatiotemporal}
Z.~Xia, X.~Hong, X.~Gao, X.~Feng, and G.~Zhao, ``Spatiotemporal recurrent
  convolutional networks for recognizing spontaneous micro-expressions,''
  \emph{IEEE Transactions on Multimedia}, vol.~22, no.~3, pp. 626--640, 2019.

\bibitem{wei2022learning}
J.~Wei, G.~Lu, J.~Yan, and Y.~Zong, ``Learning two groups of discriminative
  features for micro-expression recognition,'' \emph{Neurocomputing}, vol. 479,
  pp. 22--36, 2022.

\bibitem{xia2020learning}
B.~Xia, W.~Wang, S.~Wang, and E.~Chen, ``Learning from macro-expression: a
  micro-expression recognition framework,'' in \emph{ACM Multimedia}, 2020, pp.
  2936--2944.

\bibitem{xia2021micro}
B.~Xia and S.~Wang, ``Micro-expression recognition enhanced by macro-expression
  from spatial-temporal domain.'' in \emph{IJCAI}, 2021, pp. 1186--1193.

\bibitem{lei2021micro}
L.~Lei, T.~Chen, S.~Li, and J.~Li, ``Micro-expression recognition based on
  facial graph representation learning and facial action unit fusion,'' in
  \emph{Proceedings of the IEEE/CVF conference on computer vision and pattern
  recognition}, 2021, pp. 1571--1580.

\bibitem{wang2023temporal}
T.~Wang and L.~Shang, ``Temporal augmented contrastive learning for
  micro-expression recognition,'' \emph{Pattern Recognition Letters}, vol. 167,
  pp. 122--131, 2023.

\bibitem{see2019megc}
J.~See, M.~H. Yap, J.~Li, X.~Hong, and S.-J. Wang, ``Megc 2019--the second
  facial micro-expressions grand challenge,'' in \emph{2019 14th IEEE
  International Conference on Automatic Face \& Gesture Recognition (FG
  2019)}.\hskip 1em plus 0.5em minus 0.4em\relax IEEE, 2019, pp. 1--5.

\bibitem{zhao2021two}
S.~Zhao, H.~Tao, Y.~Zhang, T.~Xu, K.~Zhang, Z.~Hao, and E.~Chen, ``A two-stage
  3d cnn based learning method for spontaneous micro-expression recognition,''
  \emph{Neurocomputing}, vol. 448, pp. 276--289, 2021.

\bibitem{valstar2010induced}
M.~Valstar, M.~Pantic \emph{et~al.}, ``Induced disgust, happiness and surprise:
  an addition to the mmi facial expression database,'' in \emph{Proc. 3rd
  Intern. Workshop on EMOTION (satellite of LREC): Corpora for Research on
  Emotion and Affect}, vol.~10.\hskip 1em plus 0.5em minus 0.4em\relax Paris,
  France., 2010, p.~65.

\end{thebibliography}
%
%
%
\begin{IEEEbiography}[{\includegraphics[width=1in,height=1.25in,clip,keepaspectratio]{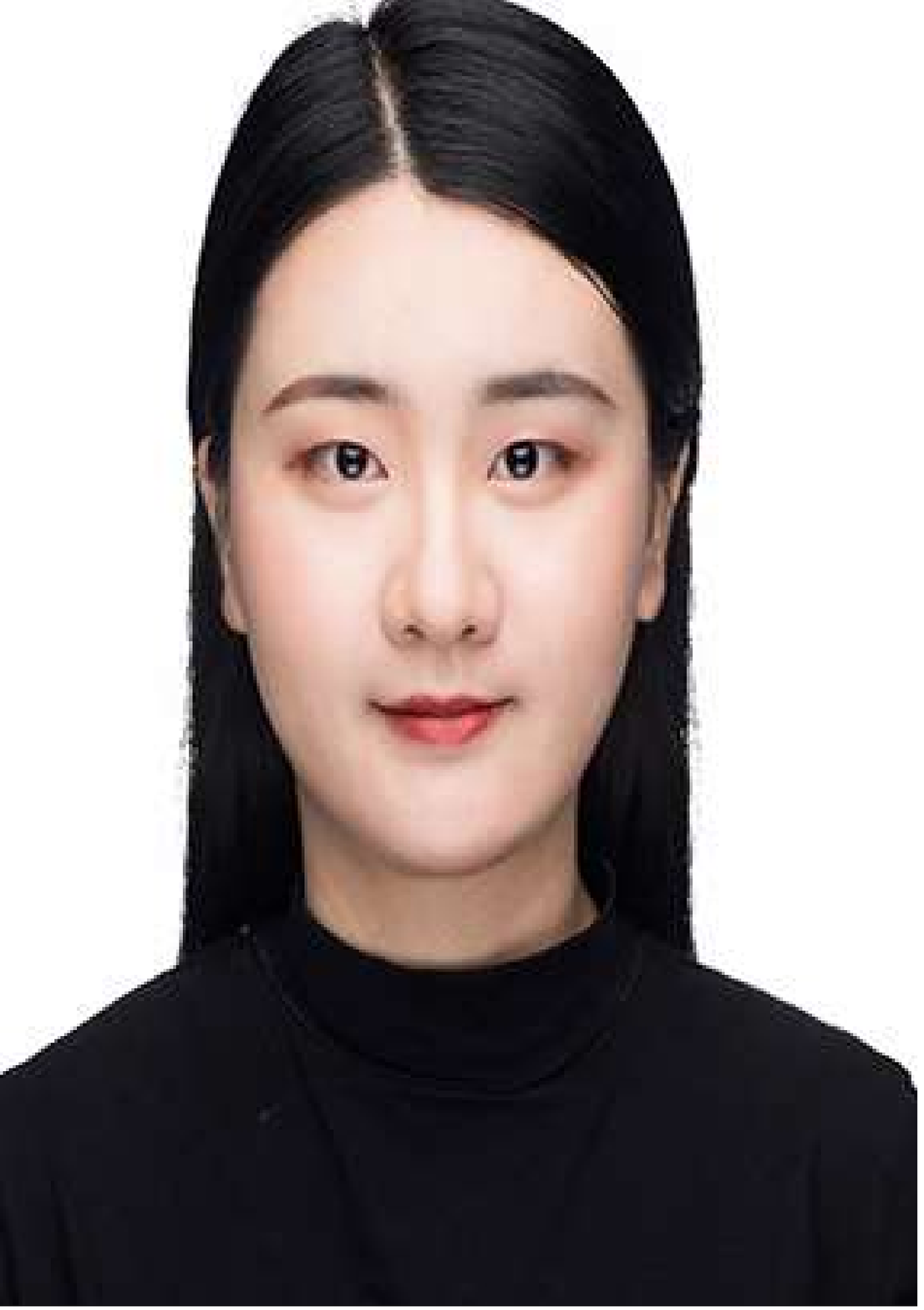}}] {Jie Zhu} received the B.S. degree from the School of Computer and Information, Hohai University, Nanjing, China, in 2018. She is currently pursuing the Ph.D. degree with the Key Laboratory of Child Development and Learning Science of Ministry of Education, Southeast University, Nanjing, China, and also with the School of Information Science and Engineering, Southeast University.

Her research interests include affective computing and computer vision.
\end{IEEEbiography}

\begin{IEEEbiography}[{\includegraphics[width=1in,height=1.25in,clip,keepaspectratio]{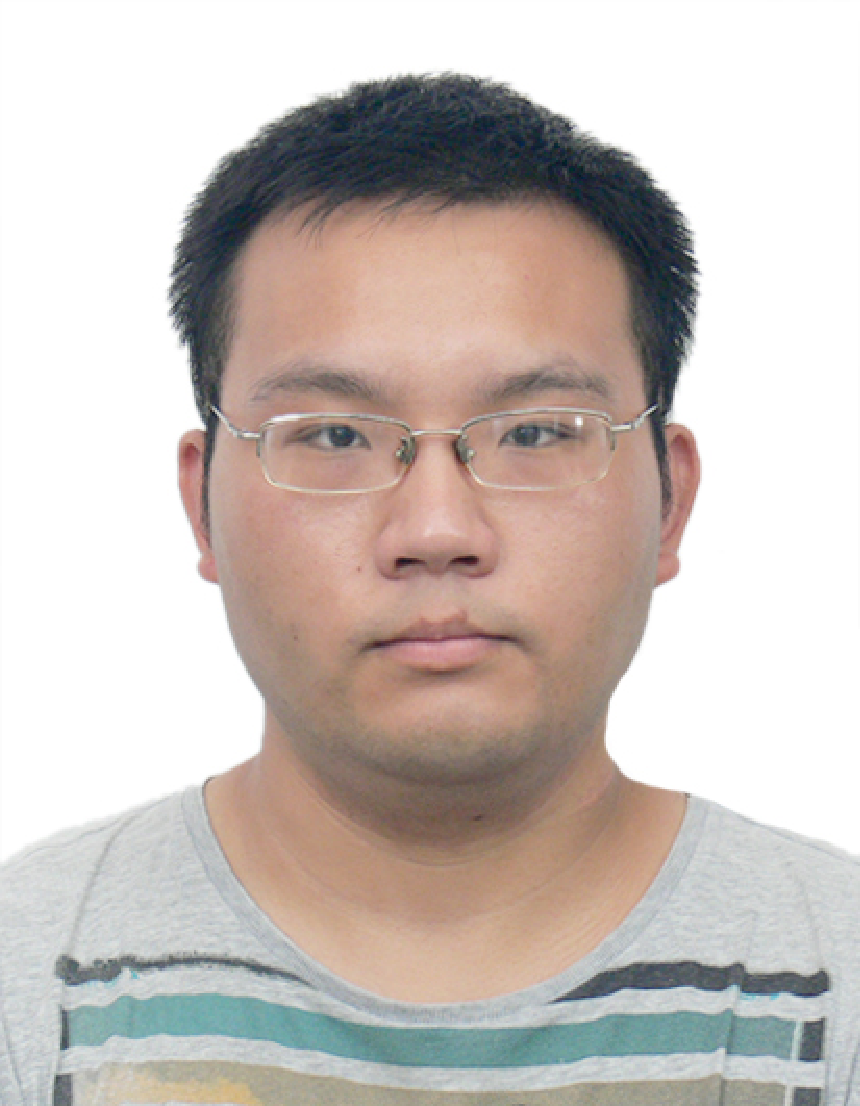}}] {Yuan Zong} (Member, IEEE) received the B.S. and M.S. degrees in electronics engineering from the Nanjing Normal University, Nanjing, China, in 2011 and 2014, respectively, and the Ph.D. degree in biomedical engineering from the Southeast University, Nanjing, China, in 2018. He is currently an Associate Professor with the Key Laboratory of Child Development and Learning Science of Ministry of Education, School of Biological Science and Medical Engineering, Southeast University. From 2016 to 2017, he was a Visiting Student with the Center for Machine Vision and Signal Analysis, University of Oulu, Oulu, Finland.

He has authored or coauthored more than 30 articles in mainstream journals and conferences, such as the IEEE T-IP, T-CYB, T-AFFC, AAAI, IJCAI, and ACM MM. His research interests include affective computing, pattern recognition, computer vision, and speech signal processing.
\end{IEEEbiography}

\begin{IEEEbiography}[{\includegraphics[width=1in,height=1.25in,clip,keepaspectratio]{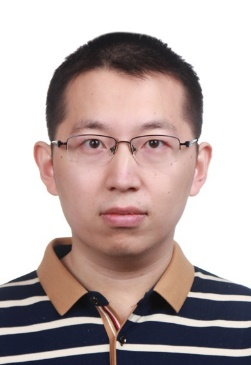}}] {Jingang Shi} received the B.S. and Ph.D. degrees from the School of Electronics and Information Engineering, Xi’an Jiaotong University, Xi’an, China. From 2017 to 2020, he was a Postdoctoral Researcher with the Center for Machine Vision and Signal Analysis, University of Oulu, Finland. Since 2020, he has been an Associate Professor with the School of Software, Xi’an Jiaotong University.

His current research interests mainly include image restoration, face analysis, and biomedical signal processing.
\end{IEEEbiography}

\begin{IEEEbiography}[{\includegraphics[width=1in,height=1.25in,clip,keepaspectratio]{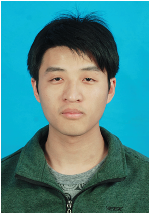}}] {Cheng Lu} received the B.S. and M.S. degrees in computer science and technology both from the Anhui University, Hefei, China, in 2013 and 2017, respectively, and the Ph.D. degree in information and communication engineering from the Southeast University, Nanjing, China, in 2023. He is currently a Zhishan Postdoctoral Researcher with the Key Laboratory of Child Development and Learning Science of Ministry of Education, School of Biological Science and Medical Engineering, Southeast University.

His research interests include affective computing, computer vision, and speech signal processing.
\end{IEEEbiography}

\begin{IEEEbiography}[{\includegraphics[width=1in,height=1.25in,clip,keepaspectratio]{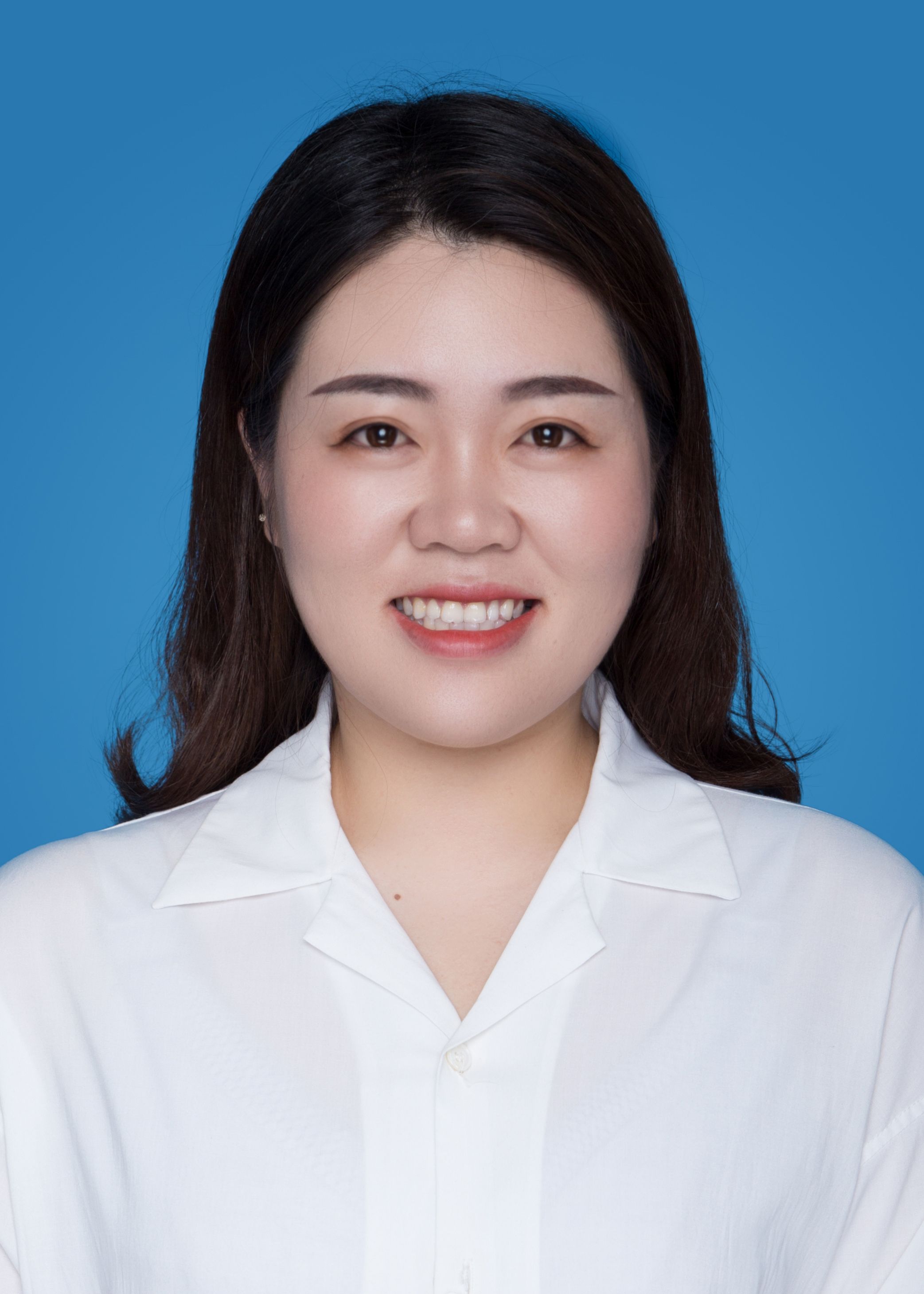}}] {Hongli Chang}  received the B.S. and M.S. degrees in electronics engineering both from Shandong Normal University, Jinan, China, in 2016 and 2019, respectively. She is currently pursuing a Ph.D. degree in information and communication engineering with Southeast University, Nanjing, China.

Her research interests include brain-computer interface, affective computing, machine learning, pattern recognition, and depression diagnosis.
\end{IEEEbiography}

\begin{IEEEbiography}[{\includegraphics[width=1in,height=1.25in,clip,keepaspectratio]{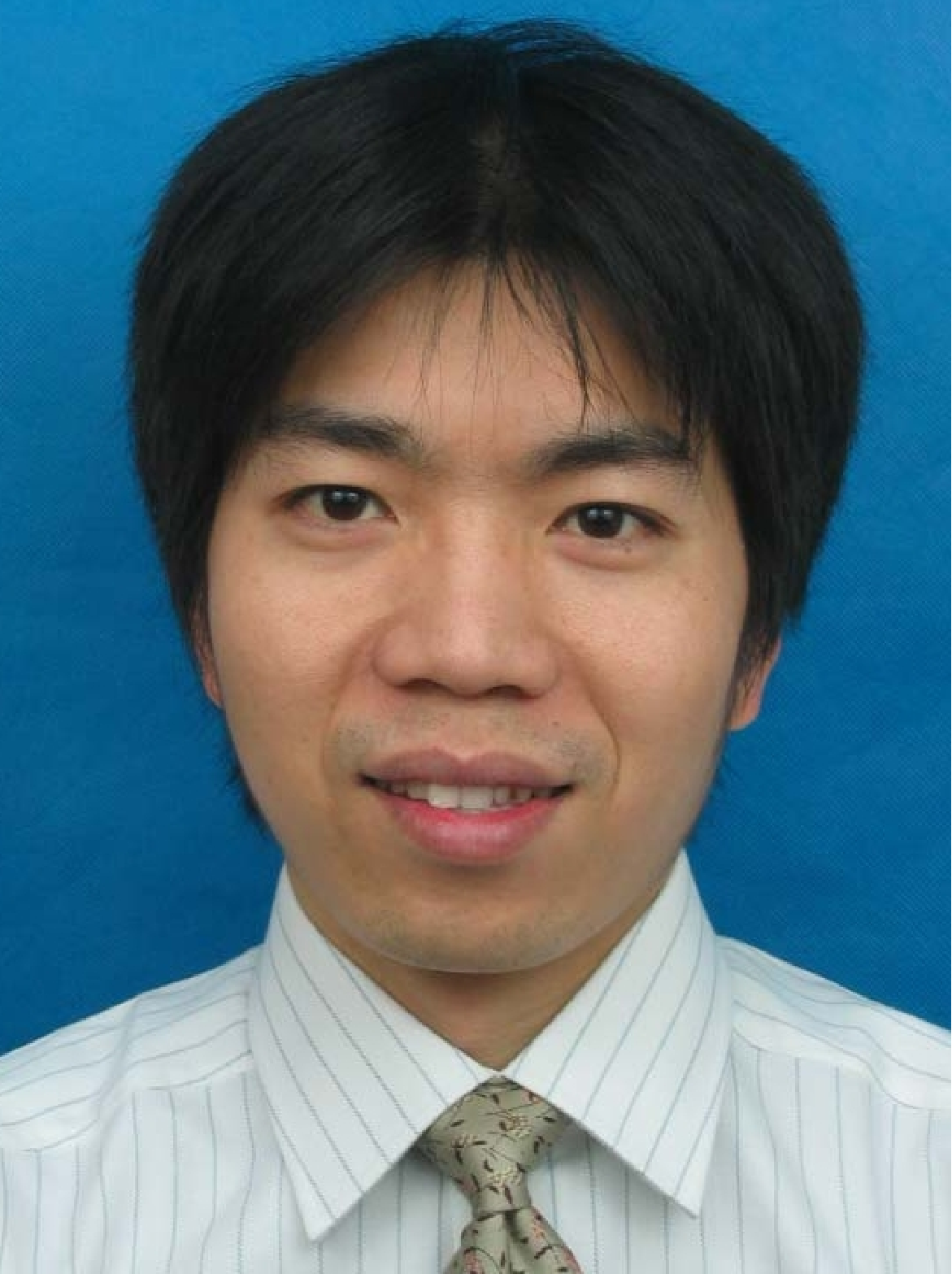}}] {Wenming Zheng} (Senior Member, IEEE) received the B.S. degree in computer science from the Fuzhou University, Fuzhou, China, in 1997, the M.S. degree in computer science from the Huaqiao University, Quanzhou, China, in 2001, and the Ph.D. degree in signal processing from the Southeast University, Nanjing, China, in 2004. He is currently a Professor with the the Key Laboratory of Child Development and Learning Science of Ministry of Education, School of Biological Science and Medical Engineering, Southeast University. He has been elected as a Fellow of IET since 2022.

His research interests include affective computing, pattern recognition, machine learning and computer vision. He is currently an Associated Editor for
the IEEE TRANSACTIONS ON AFFECTIVE COMPUTING and the Editorial Board Member of \textit{The Visual Computer}.
\end{IEEEbiography}
%

%
%
%




\end{document}